%% file: main.tex
\documentclass{article}

\PassOptionsToPackage{numbers}{natbib}

\usepackage[preprint]{neurips_2022}

\usepackage{color-edits}
\addauthor{sw}{blue}

\input{math_commands.tex}

\usepackage[utf8]{inputenc} 
\usepackage[T1]{fontenc}    
\usepackage{hyperref}       
\usepackage{url}            
\usepackage{booktabs}       
\usepackage{amsfonts}       
\usepackage{nicefrac}       
\usepackage{microtype}      
\usepackage{xcolor}         

\usepackage{algorithmic}
\usepackage[ruled]{algorithm2e}
\usepackage{apxproof}
\usepackage{bbm}
\usepackage{graphicx}
\usepackage{multirow}
\usepackage{xspace}
\usepackage{enumitem}

\title{Private Synthetic Data with Hierarchical Structure}

\author{%
  Terrance Liu \\
  Carnegie Mellon University\\
  Pittsburgh, PA 15213 \\
  \texttt{terrancl@cs.cmu.edu} \\
  \And
  Zhiwei Steven Wu\\
  Carnegie Mellon University\\
  Pittsburgh, PA 15213 \\
  \texttt{zstevenwu@cmu.edu} \\
}

\newcommand{\hpd}{\mbox{{\sf HPD}}\xspace}
\newcommand{\generative}{\mbox{{\sf HPD-Gen}}\xspace}
\newcommand{\fixed}{\mbox{{\sf HPD-Fixed}}\xspace}

\newcommand{\pep}{\mbox{{\sf PEP}}\xspace}
\newcommand{\mwem}{\mbox{{\sf MWEM}}\xspace}
\newcommand{\gem}{\mbox{{\sf GEM}}\xspace}
\newcommand{\rapsoftmax}{\mbox{{\sf RAP\textsuperscript{softmax}}}\xspace}

\newcommand{\mwemupdate}{\mbox{{\sf MWEM-Update}}\xspace}
\newcommand{\hpdupdate}{\mbox{{\sf HPD-Update}}\xspace}
\newcommand{\randomize}{\mbox{{\sf Randomize}}\xspace}
\newcommand{\ema}{\mbox{{\sf EMA}}\xspace}

\newcommand{\linear}{\mbox{linear query}\xspace}
\newcommand{\linears}{\mbox{linear queries}\xspace}

\newcommand{\adaptive}{\mbox{{\sf Adaptive Measurements}}\xspace}

\begin{document}

\newtheoremrep{inequality}{Inequality}
\newtheoremrep{theorem}{Theorem}
\newtheoremrep{lemma}{Lemma}
\newtheoremrep{definition}{Definition}
\newtheoremrep{proposition}{Proposition}

\maketitle

\begin{abstract}

We study the problem of differentially private synthetic data generation for hierarchical datasets in which individual data points are grouped together (e.g., people within households). In particular, to measure the similarity between the synthetic dataset and the underlying private one, we frame our objective under the problem of private query release, generating a synthetic dataset that preserves answers for some collection of queries (i.e., statistics like mean aggregate counts). However, while the application of private synthetic data to the problem of query release has been well studied, such research is restricted to non-hierarchical data domains, raising the initial question---what queries are important when considering data of this form? Moreover, it has not yet been established how one can generate synthetic data at both the \textit{group} and \textit{individual}-level while capturing such statistics. In light of these challenges, we first formalize the problem of \textit{hierarchical query release}, in which the goal is to release a collection of statistics for some hierarchical dataset. Specifically, we provide a general set of statistical queries that captures relationships between attributes at both the \textit{group} and \textit{individual}-level. Subsequently, we introduce private synthetic data algorithms for hierarchical query release and evaluate them on hierarchical datasets derived from the American Community Survey and Allegheny Family Screening Tool data. Finally, we look to the American Community Survey, whose inherent hierarchical structure gives rise to another set of domain-specific queries that we run experiments with.

\end{abstract}

\input{docs/introduction}

\input{docs/domain}

\input{docs/queries}

\input{docs/model}

\input{docs/experiments}

\input{docs/acs_case_study}

\input{docs/conclusion}



\bibliography{main}
\bibliographystyle{plainnat}

\input{docs/appendix}

\end{document}

%% file: math_commands.tex
\usepackage{amsmath,amsfonts,bm}

\def\eqref#1{equation~\ref{#1}}

\def\1{\bm{1}}

\def\rvx{{\mathbf{x}}}

\def\rvz{{\mathbf{z}}}

\DeclareMathAlphabet{\mathsfit}{\encodingdefault}{\sfdefault}{m}{sl}
\SetMathAlphabet{\mathsfit}{bold}{\encodingdefault}{\sfdefault}{bx}{n}

\def\gD{{\mathcal{D}}}

\def\gG{{\mathcal{G}}}

\def\gI{{\mathcal{I}}}

\def\gL{{\mathcal{L}}}
\def\gM{{\mathcal{M}}}
\def\gN{{\mathcal{N}}}

\def\gQ{{\mathcal{Q}}}

\def\gX{{\mathcal{X}}}
\def\gY{{\mathcal{Y}}}

\def\sR{{\mathbb{R}}}

\DeclareMathOperator*{\argmin}{arg\,min}

\newcommand{\pp}[1]{\left(#1\right)}
\newcommand{\cc}[1]{\left\{#1\right\}}
\newcommand{\anglebrack}[1]{\left\langle#1\right\rangle}
\newcommand{\sbrack}[1]{\left[#1\right]}

\newcommand{\norm}[1]{\left\lVert #1 \right\rVert}

\newcommand{\vect}[1]{\mathbf{#1}}

%% file: docs/introduction.tex
\section{Introduction}

Differential privacy \citep{DworkMNS06} provides rigorous privacy guarantees that center around limiting the influence of any individual data point. As a result, organizations have increasingly adopted differential privacy to release information while protecting individuals' sensitivite data. The 2020 U.S. Decennial Census, for example, serves as one of the most prominent deployments of differential privacy in recent years \citep{Abowd18}. In this work, we study \textit{private synthetic data generation} for the purpose of preserving a large collection of summary statistics. This task, which is otherwise known as \textit{private query release}, is one of most fundamental problems in differential privacy and remains a key objective for many organizations like the U.S. Census Bureau. For example, recent work \citep{hardt2010simple, gaboardi2014dual, vietri2020new, liu2021leveraging, aydore2021differentially, liu2021iterative, mckenna2022aim} has demonstrated the effectiveness of generating private synthetic data to solve query release; and after announcing plans to incorporate differential privacy into the American Community Survey (ACS) \citep{jarmin2025acs}, the U.S. Census Bureau declared (albeit informally) that it intended to replace the public ACS release with fully synthetic data \citep{rodriguez2021synacs}.

While private synthetic data generation has shown promise with respect to query release, prior work has not studied the setting in which the data exhibits hierarchical structure (i.e., individual data points that are naturally grouped together). In particular, the synthetic data generation process for records at both the \textit{group} and \textit{indivdiual}-level has not yet been established in this problem domain. Moreover, although studies based on hierarchical data often look at relationships between individuals across groups (e.g., trends observed between domestic partners), queries in the traditional setting do not describe such statistics. As a result, social scientists have criticized the U.S. Census Bureau's objective to replace the ACS with synthetic data, arguing that existing methods are unsuitable since they can only capture statistical relationships at the individual-level \citep{ipums_acssynthetic}.

\textbf{Our Contributions.} As alluded to by such critiques, private synthetic data generation remains an open problem for hierarchical data within the context of query release. Therefore, the objective of this work is to explore this problem and introduce differentially private algorithms for synthetic data with hierarchical structure. In pursuit of this goal, we make the following contributions:
\begin{enumerate}[leftmargin=*]
    \setlength\itemsep{0em}
    \item We initiate the study of \textit{hierarchical query release}, formulating the problem in which the data domain has two levels---denoted as \textit{group} and \textit{individual}\footnote{In contrast, previous works only consider data domains containing the latter.}---in its data hierarchy.
    
    \item We construct a set of queries that describe statistical relationships across \textit{group} and \textit{individual}-level attributes, where our goal is to capture how attributes interact both within and between each level.
    
    \item While our formulation of hierarchical data domains naturally lends itself to running synthetic data generation algorithms such as \mwem \citep{hardt2010simple}, these algorithms are computationally intractable. Therefore, inspired by methods from \citet{liu2021iterative}, we present a practical approach for hierarchical query release that models the data as separate product distributions over \textit{group} and \textit{individual}-level attributes. Using this approach, we introduce two methods, \fixed and \generative, which we empirically evaluate on hierarchical datasets that we construct from the American Community Survey (ACS)\footnote{To allow researchers to reproduce our evaluation datasets, code for preprocessing ACS data derived from IPUMS USA \citep{ruggles2021ipums} will be made available online.} and Allegheny Family Screening Tool (AFST) data.
    
    \item We use the ACS to formulate another set of queries meant to capture relationships \textit{between people} within households (e.g., the likelihood that an individual with a graduate degree is \textit{married to} someone who also has a graduate degree). We run additional experiments in this setting.
\end{enumerate}

\textbf{Related Work.} Query release remains as one of the most fundamental problems in differential privacy \citep{BlumLR08}, and over the years, synthetic data generation has become a well studied approach, both from theoretical \citep{RothR10, hardt2010multiplicative, hardt2010simple, GRU} and practical \citep{gaboardi2014dual, vietri2020new, aydore2021differentially} perspectives. In particular, \citet{liu2021iterative} unify iterative approaches to synthetic data generation for private query release under their framework \adaptive and concurrently with \citet{aydore2021differentially} propose methods that take advantage of gradient-based optimization. Moreover, \citet{liu2021iterative} demonstrate that the performance of their iterative algorithms can be improved significantly when optimizing over marginal queries, where the $\ell_2$ sensitivity can be bounded for an entire workload of queries. In later work, \citet{mckenna2022aim} also explore this case, presenting an iterative procedure that achieves strong performance while removing the burden of hyperparameter tuning.

Despite this long line of research applying synthetic data to private query release, prior works---including those that have used the ACS itself for empirical evaluation \citep{liu2021leveraging, liu2021iterative}---are limited to the non-hierarchical setting. In addition, we note that there exist other works that do not tackle this specific problem: (1) those studying private query release using "data-independent" mechanisms \citep{mm, mckenna2018optimizing, mckenna2019graphical, NTZ, ENU20} and (2) those studying synthetic (tabular) data generation for purposes other than query release \citep{rmsprop_DPGAN, yoon2018pategan, xu2019modeling, beaulieu2019privacy, neunhoeffer2020private, rosenblatt2020differentially}. However, such works also do not consider hierarchical structure.

%% file: docs/domain.tex
\section{Formulation of Hierarchical Data Domains}\label{sec:domain}


We consider data with some hierarchical structure, in which a dataset can first be partitioned into different groups that can then further be divided into individual rows. For example, one can form a hierarchical dataset from census surveys by grouping individuals into their respective households. We assume that each group contains at most $M > 1$ rows and has $k_G$ features $\gG = \pp{\gG_{1} \times \ldots \times \gG_{k_G}}$. Similarly, we assume that each individual row belongs to some data domain of $k_I$ features $\gI = \pp{\gI_{1} \times \ldots \times \gI_{k_I}}$. For example, a census survey may record whether households reside in an urban or rural area ($\gG_1 = \textsc{Urban/Rural}$) and what the sex and race ($\gI_1 = \textsc{Sex}, \gI_2 = \textsc{Race}$) are of individuals in each household. Putting these attribute domains together, we then have a hierarchical data universe over all possible groups of size no more than $M$:\footnote{In the traditional (non-hierarchical) query release setting, there instead exists some private dataset $D \in \gX$ with $k$ discrete attributes, such that the data domain $\gX = \gX_1 \times \ldots \times \gX_k$.}
\begin{equation}\label{eq:data_domain}
    \gX = \gG \times \tilde{\gI}^M,
\end{equation}
where $\tilde{\gI} = \gI \times \perp$ and $\perp$ represents an empty set of features in $\gI$ (i.e., one of the $M$ possible individual rows in the group does not exist). In other words, every element in $\gX$ corresponds to a possible group (including the individual rows it contains) in the data universe. Letting the \textit{group}-level domain size $d_G = \prod_{i=1}^{k_G} |\gG_i|$ and \textit{individual}-level domain size $d_I = \prod_{i=1}^{k_I} |\gI_i|$, the overall domain size can be written as $d = d_G \pp{d_I + 1}^M$. 

Given this formulation of $\gX$, our goal then is to study the problem of answering a collection of queries $Q$ about some dataset $D \in \gX$ while satisfying differential privacy.
\begin{definition}[Differential Privacy \citep{DworkMNS06}]\label{def:dp}
A randomized mechanism $\gM:\gX^n \rightarrow \sR$ is $(\varepsilon,\delta)$-differentially privacy, if for all neighboring datasets $D, D'$ (i.e., differing on a single person), and all measurable subsets $S\subseteq \sR$ we have:
\begin{align*}
    P\pp{\gM(D) \in S} \leq e^{\varepsilon} P\pp{\gM(D') \in S} + \delta
\end{align*}
\end{definition}

Typical differentially private query release algorithms operate under the assumption that \textit{neighboring datasets} $D, D'$ differ on a single row where $N = |D| = |D'|$ is given as public knowledge.\footnote{This version of differential privacy that uses this notion of neighboring datasets is sometimes referred to as \textit{bounded differential privacy} \citep{Dwork06, DworkMNS06}}
In our setting, however, we have a dataset $D \in \gX$ with $N_I$ individual rows comprising $N_G$ groups of size $m \le M$. Therefore, we similarly assume that $N_G$, $N_I$, and $M$ are given (and fixed). For release of census survey data, for example, the U.S. Census Bureau may publish publicly both the total number of people in the United States and the total number of households that such people reside in. Moreover, we assume that while $D$ and $D'$ still differ on an individual row, the total number of groups and maximum group size $M$ remain the same.

%% file: docs/queries.tex
\section{Hierarchical Query Release}\label{sec:queries}

We now formulate the problem of \textit{hierarchical query release}, in which we construct a set of queries that our synthetic data algorithms will optimize over. Specifically, to construct a set of statistics describing datasets $D \in \gX$, we will focus on \linears, which we define as the following:
\begin{definition}[linear query] \label{def:linear_query}
Given a dataset $D$ and predicate function $\phi:\gX \rightarrow \sR$, a \linear is defined as
\begin{align*}
    q_\phi(D) = \sum_{x \in D} \phi(x)
\end{align*}
\end{definition}

For example, the number of males in some dataset can be represented as a \linear with a predicate function $\phi(x) = \mathbbm{1}\cc{x \textrm{ is male}}$. Similar to previous work, we will also normalize query answers with respect to a fixed constant $N$, where $N$ denotes some overall count such as the size of the dataset $D$ (e.g. \% \textit{males} = \# \textit{males} / \# \textit{people}).\footnote{In some prior works, a \linear is defined with the normalizing constant included (where $N = |D|$).} 

In the case where $\phi$ is an indicator function, the output of some \linear $q_{\phi}(D)$ can be interpreted as the count of elements $x \in D$ satisfying a (boolean) condition defined by $\phi$ (e.g., \textit{is male}). Going forward, we will refer to such predicates as \textbf{singleton predicate} functions, where we restrict $\phi$ to condition over a single attribute in either $\gG$ or $\gI$. We can then form \linears over $k$ distinct attributes by combining singleton predicates into a \textit{$k$-way predicate} function, which we define as
\begin{definition}[$k$-way predicate] \label{def:kway_predicate}
Given some set of $k$ singleton predicate functions $\phi_1, \phi_2, \ldots, \phi_k$, we define a $k$-way predicate function as the product $\phi(x) = \prod_{i=1}^k \phi_i(x)$.
\end{definition}
In other words, a $k$-way predicate function outputs some logical conjunction over boolean conditions defined for $k$ distinct attributes in $\gX$ (e.g., whether a person is both \textit{male} and \textit{30 years old}).

In the traditional query release setting, it suffices to construct a set of queries counting the number of (individual) rows satisfying \linears defined by $k$-way predicates. In contrast, given the hierarchical structure of $\gX$, our goal then is to output counts at both the \textit{group} and \textit{individual}-level. Consequently, we have two classes of query types: \textbf{(1)} \linears at the \textbf{\textit{group}}-level $\gQ_G$ (i.e., \% of households) and \textbf{(2)} \linears at the \textbf{\textit{individual}}-level $\gQ_I$ (i.e., \% of people).

\textbf{Group-level queries.} Given some dataset $D \in \gX$, we have that each element $x \in D$ corresponds to a group. Therefore, we can describe write down \textit{group-level} counting queries using Defintion \ref{def:linear_query}, where $\phi: \gX \rightarrow \cc{0, 1}$ is some $k$-way predicate function.

\textbf{Individual-level queries.} When counting individual rows, each group $x \in \gX$ can contribute up to $M$ rows to the total count (e.g., a household can contain up to $M$ males). Therefore, defining such queries using Definition \ref{def:linear_query} requires that $\phi:\gX \rightarrow \cc{0, 1, \ldots, M}$. However, we can convert hierarchical datasets from a collection of groups $D \in \gX$ to a collection of individuals $D \in \gY$, where
\begin{equation}\label{eq:data_domain_alt}
    \gY = \gI \times \gG \times \tilde{\gI}^{M-1}
\end{equation}
Unpacking Equation \ref{eq:data_domain_alt}, we have that $\gI$ describes the features for some individual row $x$, $\gG$ the group to which $x$ belongs, and $\tilde{\gI}^{M-1}$ the features of the remaining group members. Now, we can also write down these \linears using $k$-way predicates, where $\phi:\gY \rightarrow \cc{0, 1}$.
 
\input{tables/query_types}
 
We summarize in Table \ref{tab:query_types} the singleton predicates for both \textit{group} and \textit{individual}-level queries that we consider. Given some set $S_G$ of \textit{group}-level attributes and set $S_I$ of \textit{individual}-level attributes where $k = |S_G| + |S_I|$, we construct a set of \textit{group} and \textit{individual}-level queries by combining singleton predicates for each attribute in $S_G \cup S_I$. For example, given the \textit{group}-level attribute \textsc{Urban/rural} and \textit{individual}-level attributes \textsc{Sex} and \textsc{Race}, we have queries of the following form:\footnote{i.e., \linears where $\phi(x) = \phi_{[\textsc{Urban/rural = URBAN}]}(x) \times \phi_{[\textsc{Sex = FEMALE}]}(x) \times \phi_{[\textsc{Race = WHITE}]}(x)$.}
\begin{itemize}
    \setlength\itemsep{0em}
    \item ($\gQ_G$) What proportion of \textit{households} are located in an \textbf{urban} area and contain \textit{at least one individual} who is \textbf{female} and \textbf{white}?
    \item ($\gQ_I$) What proportion of \textit{individuals} reside in a household located in an \textbf{urban} area and are \textbf{female} and \textbf{white}? 
\end{itemize}

Finally, for differentially private algorithms, it is necessary to derive the $\ell_p$-sensitivity, which captures the effect of changing an individual in the dataset for each function (or query).
\begin{definition}[$\ell_p$-sensitivity]\label{def:sensitivity}
The $\ell_p$-sensitivity of a function $f:\gX^*\rightarrow \sR^k$ is
\begin{align*}
    \Delta_p f = \max_{\text{neighboring } D,D'} \| f(D) - f(D')\|_p
\end{align*}
\end{definition}
Because $\gQ_G$ and $\gQ_I$ are comprised of \linears with $|\phi(x)| \le 1$ and are normalized by constants $N = N_G$ and $N = N_I$ respectively, the $\ell_1$-sensitivities of queries in $\gQ_G$ and $\gQ_I$ are $\frac{1}{N_G}$ and $\frac{1}{N_I}$.

%% file: tables/query_types.tex
\begin{table}[!tb]
\caption{We can distinguish between singleton predicates $\phi$ by reducing them to (1) whether counts are made at the \textit{group} or \textit{individual}-level and (2) the domain of the attribute that $\phi$ specifies over ($\gG$ or $\gI$). In this table, given some attribute $\gG_i$ or $\gI_i$, we describe the corresponding predicate for each combination of (1) and (2) where $y$ is some target value (e.g, $y = \textsc{Male}$ for $\gI_i = \textsc{Sex}$).}
\begin{center}
\renewcommand{\arraystretch}{1.3}
\begin{tabular}{c c || c c}
\toprule
Query Type & Attr. Domain & Predicate & Example \\
\midrule
\multirow{4}{*}{$\gQ_G$} & \multirow{2}{*}{$\gG$} & \multirow{2}{*}{attribute $\gG_i = y$} & What proportion of \textit{households} \\
& & & reside in \textbf{rural} areas? \\ \cmidrule(lr){2-4}
& \multirow{2}{*}{$\gI$} & contains individual row & What proportion of \textit{households} \\
& & with attribute $\gI_i = y$ & contain (at least) \textbf{one male} individual?\\
\hline
\multirow{4}{*}{$\gQ_I$} 
& \multirow{2}{*}{$\gG$} & belongs to a group & What proportion of \textit{individuals} \\
& & with attribute $\gG_i = y$ & live in \textbf{rural} households? \\ \cmidrule(lr){2-4}
& \multirow{2}{*}{$\gI$} & \multirow{2}{*}{attribute $\gI_i = y$} & What proportion of \textit{individuals} \\
& & & are \textbf{male}? \\
\bottomrule
\end{tabular}
\label{tab:query_types}
\end{center}
\end{table}

%% file: docs/model.tex
\section{Synthetic data generation for hierarchical data}\label{sec:modeling}
 
Having formulated the problem of hierarchical query release, we next introduce a general approach to answering queries privately via synthetic data generation. Given some family of data distributions $\gD$, our goal then is to find some synthetic dataset $D' \in \gD$ that solves the optimization problem
\begin{align*}
    \min_{\hat{D} \in \gD} \max_{q \in \gQ} | q(D) - q(\hat{D})|
\end{align*}

One natural approach is construct a probability distribution over all elements in $\gX$, and while methods using this form of $\gD$ have been shown to perform well \citep{hardt2010multiplicative, hardt2010simple, ullman2015private, gaboardi2014dual, liu2021leveraging, liu2021iterative}, they are computationally intractable in most settings given that the size of $\gX$ grows exponentially with the number of attributes. Inspired by methods introduced in \citet{liu2021iterative}, we demonstrate that an alternative solution is to represent $\gD$ as a mixture of product distributions over each attribute in $\gI$ and $\gG$. In this way, the computational requirements for methods operating on $\gD$ scale only \textit{linearly} with respect to $M$ and the number of attributes $k_G$ and $k_I$.

\textbf{Adaptive Measurements.} We consider algorithms under the \adaptive framework \citep{liu2021iterative}, which we provide a brief overview for. Having observed that many synthetic data generation algorithms for private query release share similar iterative procedures, \citet{liu2021iterative} present this unifying framework for such algorithms that can broken down into the following steps at each round $t$: \textbf{(1)} sample privately some query with high error, \textbf{(2)} measure the answer to this query using a differentially private mechanism, and \textbf{(3)} optimize some loss function with respect to measurements from all rounds $1, \ldots, t$. Moreover, they show that the formulation of algorithms under this framework can then be reduced to selecting $\gD$ and finding a corresponding loss function $\gL$.

\input{algos/adaptive}

We restate \adaptive in Algorithm \ref{alg:adaptive} using the \textit{exponential} and \textit{Gaussian} mechanisms for the sample and measure steps respectively. We also note that in our algorithms, we will run these mechanisms under the assumption that all queries $q \in \gQ$ have $\ell_1$-sensitivity equal to $\max_{q \in \gQ} \Delta_1\pp{q}$. For example, we have a sensitivity of $\frac{1}{N_G}$ for $\gQ = \gQ_G \cup \gQ_I$ (since $N_G \ge N_I$). Having bounded the sensitivity of the queries, we can now preserve the privacy guarantees of \adaptive for our hierarchical setting (we provide a proof in Appendix \ref{appx:adaptive}).
\begin{theorem}
For all $\delta > 0$, there exist parameters $\varepsilon_0, \varepsilon_1$ such that when run with the exponential and Gaussian mechanisms using $\varepsilon_0$ and $\varepsilon_1$ respectively, algorithms under the \adaptive framework satisfy $\pp{\varepsilon, \delta}$-differential privacy.
\label{thm:privacy}
\end{theorem}

\subsection{Hierarchical product distributions (HPD)}

Our goal is derive a probability distribution from which we sample groups (attributes $\gG$) and the individual rows (attributes $\gI$) contained in each group. Let $P_{\textrm{\#}}$ be a distribution over the possible number of individuals in each group, and let $P_{\gG}$ be the distribution from which we sample attributes $\gG$. In addition, for $i \in \cc{1, \ldots, M}$, let $P_{\gI, i}$ denote the distribution over attributes $\gI$ such that for each group, the $i^{th}$ individual row is sampled from $P_{\gI, i}$. In Algorithm \ref{alg:hpd_sampling}, we present our two-stage sampling procedure, which can be summarized as the following: (1) sample the \textit{group}-level attributes and group size $m$, and (2) sample $m$ sets of individual-level attributes. 

\input{algos/hpd_sampling}

\textbf{Parameterization.}
Inspired by \citet{liu2021iterative}, we choose to parameterize $P_{\gG}$ and $P_{\gI, i}$ as a uniform mixture of $K$ product distributions over each \textit{group} and \textit{individual}-level attribute $\gG_i$ and $\gI_j$. Concretely, our parameterization means that for each of the $K$ product distributions, we maintain a distribution over each attribute separately. For example, in the ACS dataset, the attribute $\gI_j = \textsc{Sex}$ takes on the values \textsc{Male} and \textsc{Female}, meaning that we maintain a $2$-dimensional vector corresponding to probabilities that an individual is male or female. 

\textit{Mixture of product distributions.}
One can interpret $K$ as some tunable parameter that increases the expressiveness of our probability distribution $P$. When $K=1$, we are limited to a single product distribution, meaning that one must assume in the private dataset that for all groups and the individual rows, attributes are independent. On the opposite side of the spectrum, suppose we let $K = N_G$. Then any dataset with size $N_G$ groups can be perfectly captured since each row $x \in D$ can be thought of as a product distribution over point masses. In non-hierarchical settings, \citet{liu2021iterative} show that $K$ can be much smaller than $|D|$ and still achieve strong performance.

\textit{Linear number of parameters.}
We note that the number of parameters grows linearly with the number of attributes $k_G$ and $k_I$. In particular, $P_{\gG}$ and $P_{\gI, i}$ require $K \times d'_G$ and $K \times d'_I$ parameters respectively, where $d'_G = \sum_{i=1}^{k_G} |\gG_i|$ and $d'_I = \sum_{j=1}^{k_I} |\gI_j|$.

\textit{Differentiable query answers.}
An advantage of this representation is that mean aggregate queries can be calculated using basic operations over the probabilities for each attribute $\gG_i$ or $\gI_j$. Therefore, this formulation makes constructing differentiable functions for each predicate in Table \ref{tab:query_types} convenient. For example, to calculate that the likelihood $P\sbrack{\gI_{\textsc{Sex}, j} = \textsc{Male}}$ that an individual generated from $P_{\gI, j}$ is \textit{male}, one can simply look up the probability in the $2$-dimensional vector corresponding to the attribute $\gI_j = \textsc{Sex}$. Similarly, to calculate if there is \textit{at least one male}, it suffices to calculate the probability that a group (given some size $m$) has \textit{no males}: $1 - \prod_{j=1}^m \pp{1 - P\sbrack{\gI_{\textsc{Sex}, j} = \textsc{Male}}}$.

\subsection{Methods}

For notation purposes, we let $\vect{P}^{\gG} \in \mathcal{R}^{K \times d'_G}$ be some matrix of probability values such that $k^{th}$ row $\vect{P}^{\gI, j}_k$ is the concatenation of the $k_g$ probability vectors for attributes in $\gG$. We define $\vect{P}^{\gI} \in \mathcal{R}^{K \times d'_I}$ and $\vect{P}^{\textrm{\#}} \in \mathcal{R}^{K \times 1}$ similarly for \textit{individual}-level attributes and the group size $m$. In Figure \ref{fig:P_example}, we provide an example of a row in $\vect{P}^{\gI}$.

\input{figures/P_example}

We now describe two methods that are inspired by the family distributions $\gD$ used in \rapsoftmax and \gem \citep{liu2021iterative}. The first of which maintains a fixed distribution in which the parameters of our model are exactly the probability vectors making up our $K$ product distributions. The second of which employs generative modeling by using some neural network architecture that takes as input randomly sampled Gaussian noise $\vect{z}$. Note that in both cases, we use the softmax function to ensure that these methods output probabilities while still maintaining differentiability.  

\textbf{Fixed distribution modeling (\fixed).} 
Our first method defines a "model" whose parameters $\theta^{\gG}$, $\theta^{\gI_{i}}$, and $\theta^{\textrm{\#}}$ are simply the matrices $\vect{P}^{\gG}$, $\vect{P}^{\gI_{i}}$, and $\vect{P}^{\textrm{\#}}$ respectively.

\textbf{Generative modeling (\generative).} 
Alternatively, we propose using a multi-headed neural network $F$ to generate rows from Gaussian noise $\rvz \sim \gN(0, I_{K})$. In this case, the output of each head, which we denote by $F^{\gG}(\rvz)$, $F^{\gI_{i}}(\rvz)$, and $F^{\textrm{\#}}(\rvz)$, corresponds to the $\vect{P}^{\gG}$, $\vect{P}^{\gI_{i}}$, and $\vect{P}^{\textrm{\#}}$ respectively.



\textbf{Loss function.}
Following the \adaptive framework \citep{liu2021iterative}, at each round $t$, we have a set of selected queries $\tilde{Q}_{1:t} = \cc{\tilde{q}_1, \ldots, \tilde{q}_t}$ and their noisy measurements $\tilde{M}_{1:t} = \cc{\tilde{m}_1, \ldots, \tilde{m}_t}$. Suppose we have some set of probability matrices $\vect{P}^{\gG}$, $\vect{P}^{\gI_{i}}$ for $i \in \cc{1, 2, \ldots, M}$, and $\vect{P}^{\textrm{\#}}$ that are obtained via either the fixed distribution or generative modeling methods outlined above. Abusing notation, we use $\vect{P}$ to denote the collection of these matrices. Then at each round $t$, we optimize the following differentiable loss function via stochastic gradient descent:
\begin{equation}\label{eq:hpd_loss}
    \gL^{\hpd} \pp{\vect{P}, \tilde{Q}_{1:t}, \tilde{M}_{1:t}} = 
    \sum_{i=1}^{t} \norm{\tilde{m}_i - f_{\tilde{q}_i} \pp{\vect{P}} }_2
\end{equation}
where for query $q$, there exists a differentiable function $f_q$ that maps from $\vect{P}^{\gG}$, $\vect{P}^{\gI_{i}}$, and $\vect{P}^{\textrm{\#}}$ to the corresponding query answer. Referring to the query types in Table \ref{tab:query_types}, we have that 
\begin{equation*}
    f_{q} \pp{\vect{P}} = 
    \frac{1}{K} \sum_{k=1}^K \pp{\prod_{i \in S_G} \vect{P}^{\gG}_{ki}} 
    \pp{\sum_{m=1}^{M} \vect{P}^{\textrm{\#}}_{km} \pp{1 - \prod_{j=1}^{m}{\pp{1 - \prod_{i \in S_I} \vect{P}^{\gI_{j}}_{ki} }}}}
\end{equation*}
for \textit{group}-level hierarchical queries ($\gQ_G$) and
\begin{equation*}
    \left.
    f_{q} \pp{\vect{P}} =
    \sum_{k=1}^K 
    \pp{\prod_{i \in S_G} \vect{P}^{\gG}_{ki}} 
    \pp{\sum_{m=1}^M \vect{P}^{\textrm{\#}}_{km} \sum_{j=1}^{m} \pp{\prod_{i \in S_I} \vect{P}^{\gI_{j}}_{ki} }}
    \middle/
    \sum_{k=1}^K 
    \pp{\sum_{m=1}^M m \vect{P}^{\textrm{\#}}_{km}}
    \right.
\end{equation*}
for \textit{individual}-level hierarchical queries ($\gQ_I$).

%% file: algos/adaptive.tex
\begin{algorithm}[!tbh]
\SetAlgoLined

\textbf{Input:} Private dataset $D$, set of linear queries $\gQ$, distributional family $\gD$, loss functions $\gL$, number of iterations $T$ \\
Initialize distribution $D_0 \in \gD$ \\
 \For{$t = 1, \ldots, T$}{
  \textbf{Sample}:
  Choose $\tilde{q}_{t} \in \gQ$ using the \textit{exponential mechanism}. \\
  \textbf{Measure:} 
  Let $\tilde{a}_{t} = \tilde{q}_{t}(D) + z_{t}$ where $z_t$ is Gaussian noise (\textit{Gaussian mechanism}). \\
  \textbf{Update:}
  Let $\tilde{Q}_{t} = \bigcup_{i=1}^t \tilde{q}_{t}$ and $\tilde{A}_{t} = \bigcup_{i=1}^t \tilde{a}_{t}$. Update distribution $D$:
  \begin{align*}
    D_{t} \leftarrow \argmin_{D \in \gD} \gL \pp{D_{t-1}, \tilde{Q}_{t}, \tilde{A}_{t}}
  \end{align*}
  
 }
 Output $H\pp{\cc{D_t}_{t=0}^{T}}$  where $H$ is some function over all distributions $D_t$ (such as the average)
 \caption{Overview of \adaptive}
 \label{alg:adaptive}
\end{algorithm}

%% file: algos/hpd_sampling.tex
\begin{algorithm}[!htb]
\SetAlgoLined
\textbf{Input:} Distributions $P_{\textrm{\#}}$, $P_{\gG}$, and $P_{\gI, {1}}, \ldots, P_{\gI, {M}}$ \\
Sample group size $m ~ \sim P_{\textrm{\#}}$ \\
Sample \textit{group}-level attributes $x_G \sim P_{\gG}$ \\
\For{$i = 1, \ldots, M$} {
    Sample \textit{individual}-level attributes $x_{I,i} \sim P_{\gI, i}$ \\
}
Let $x_{I,i} = \perp$ for $i \in \cc{m+1, m+2, \ldots, M}$ \\
Output $x_G$ and $x_{I,1}, \ldots, x_{I,M}$
\caption{Sampling procedure for \hpd.}
\label{alg:hpd_sampling}
\end{algorithm}

%% file: figures/P_example.tex
\begin{figure}[!hbt]
    \centering
    \includegraphics[width=0.5\linewidth]{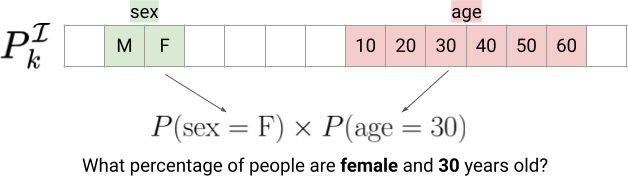}
    \caption{
    We provide an example of the $k^{th}$ row in matrix $\vect{P}^{\gI}$. As shown, the \textit{individual}-level query with the predicate for "\textit{is female}" and "\textit{is $30$}" is simply a product of two values in $\vect{P}^{\gI}$.
    }
    \label{fig:P_example}
\end{figure}

%% file: docs/experiments.tex
\section{Experiments}\label{sec:experiments}

\textbf{Methods.} 
We evaluate methods that optimize over the two distributional families mentioned in Section \ref{sec:modeling}---namely, modeling a distribution over all elements in $\gX$ (\mwem) and modeling hierarchical product distributions (\fixed and \generative).\footnote{We provide more details on these methods in Algorithms \ref{alg:mwem} and \ref{alg:hpd} in the appendix.}

\textbf{Datasets.} 
We construct datasets from the following:

\textit{American Community Survey (ACS)} \citep{ruggles2021ipums}. 
We first use the American Community Survey as our main evaluation dataset. The ACS gathers microdata annually and is widely used by various organizations to capture the socioeconomic conditions in the country. Survey responses are collected at the household level, describing the household itself ($\gG$) and the various individuals ($\gI$) residing in each household. We use data ($10$ \textit{group}-level and $11$ \textit{individual}-level attributes) from 2019 for the state of New York (\textsc{ACS NY-19}) and select households of size $M=10$ or smaller. In total, we have $N_I = 197512$ individual rows that comprise $N_G = 87453$ groups.

Running \mwem requires a domain size significantly smaller than that of \textsc{ACS NY-19}. Therefore, we create a reduced version ($4$ \textit{group}-level and $4$ \textit{individual}-level attributes) of the dataset (\textsc{ACS (red.) NY-19}), where we select households of size $M=3$ or smaller. In addition, we combine categories that attributes take on to reduce the domain size further. The domain size of \textsc{ACS (red.) NY-19} is $262080$, which is significiantly less than that of \textsc{ACS NY-19} ($d \approx 7.3 \times 10^{71}$).

\textit{Allegheny Family Screening Tool Data (AFST)} \citep{vaithianathan2017developing}. 
We run experiments using child welfare data of incidents located in Allegheny County, Pennsylvania, USA. Having been acquired from the Allegheny County Office of Children, Youth and Families (CYF), this data is used by call-screeners to make decisions about which referrals (i.e., reports of child abuse or neglect) to investigate via an AI-based tool called the Allegheny Family Screening Tool (AFST). We select a subset of columns ($5$ \textit{group}-level and $10$ \textit{individual}-level) from the source dataset where $M=10$. Our groups are the referrals themselves, which are comprised of children involved in such incidents. We will refer to this sample as \textsc{AFST}. In total, \textsc{AFST} has $N_I = 74759$ individual rows that comprise $N_G = 20448$ groups.

\textbf{Evaluation Queries.}
As shown in Section \ref{sec:queries}, given some set of \textit{group} and \textit{individual}-level attributes $S = S_G \cup S_I$, we construct \textit{group} and \textit{individual}-level queries $\gQ_G$ and $\gQ_I$. We restrict the number of attributes for each query to $|S| = 3$ and choose all such queries. This query selection scheme results in $|\gQ_G| = |\gQ_I| = 1333$ queries for \textsc{ACS (red.) NY-19}, $|\gQ_G| = |\gQ_I| = 526684$ for \textsc{ACS NY-19}, and $|\gQ_G| = |\gQ_I| = 127593$ for \textsc{AFST}.

\textbf{Results.} 
We evaluate all methods on each dataset across privacy budgets $\varepsilon \in \cc{0.125, 0.25, 0.5, 1.0}$, plotting the max error over queries $\gQ = \gQ_G \cup \gQ_I$ in Figure \ref{fig:acs_max_error}.\footnote{Plots for mean error can be found in Figures \ref{fig:acs_all_errors} of the appendix.} The methods perform reasonably well, with \fixed and \mwem outperforming \generative in all experiments. We also note that performance on \textsc{AFST} is lower due to the significantly higher sensitivity of $\gQ$ in this experiment.

\input{figures/max_error/acs}

%% file: figures/max_error/acs.tex
\begin{figure}[!tbh]
    \centering
    \includegraphics[width=\linewidth]{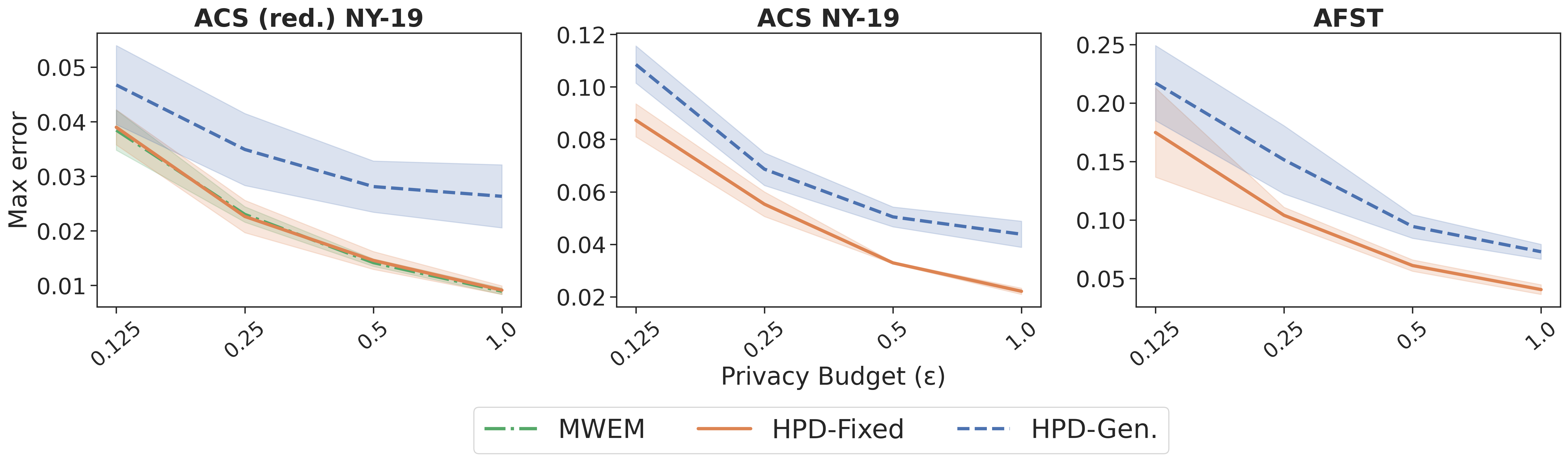}
    \caption{Max errors for \textit{group} and \textit{individual}-level hierarchical queries evaluated on ACS/ACS (red.) NY-19 and AFST where $\varepsilon \in \cc{0.125, 0.25, 0.5, 1}$ and $\delta = \frac{1}{N_I^2}$. The \textit{x-axis} uses a logarithmic scale. Results are averaged over $5$ runs, and error bars represent one standard error.
    }
    \label{fig:acs_max_error}
\end{figure}

%% file: docs/acs_case_study.tex
\section{Case study: Modeling intra-household relationships in the ACS}

Table \ref{tab:query_types} summarizes a general class of \textit{group} and \textit{individual-level} queries that are applicable to any hierarchical dataset belonging to $\gX$. However, we note that this list is non-exhaustive because of the innumerable ways in which one can ask queries about the composition of each group. Therefore, while we contend that the sets $\gQ_G$ and $\gQ_I$ serve as a good basis for capturing \textit{group} and \textit{individual-level} relationships, there could exist additional queries that are important to specific datasets or problem domains. In particular, one interesting set of statistics that may be useful to measure are those that capture relationships between individuals belonging to the same group. For example, one common analysis employed by social scientists using the ACS involves uncovering statistical trends within spousal relationships with respect to attributes like education and ethnicity. 
%

Consequently, we augment $\gQ$ for the ACS with an additional \textit{individual-level} query that takes on the form---\textit{"shares a group with another individual row with $\gI_i = y$"}---for some \textit{individual}-level attribute $\gI_i$ and target $y$. Furthermore, to focus on spousal and parent-child relationships, we restrict the dataset to only heads of the household and their spouses and children. We then let $\phi$ be predicates that identify individuals' relationships to each other (e.g., spouse/parent/child) plus one or more additional attributes (e.g., sex, age, education, etc.). Taken together, we then have \textit{intra-group-relationship} predicates such as: "has a \textbf{spouse/parent/child} that is \textbf{male}"

Combined with the predicates found in Table \ref{tab:query_types}, we can construct \textit{intra-group-relationship} queries $\gQ_R$. Suppose we have some set $S_G$ of \textit{group}-level attributes and sets $S_{I}$ and $S_{I, rel}$ of \textit{individual}-level attributes (where $S_{I, rel} \ne \emptyset$). Like in Section \ref{sec:queries} we form singleton predicates from $S_G$ and $S_{I}$ (Table \ref{tab:query_types}). However, in this case, we also add \textit{intra-group-relationship} predicates formed from attributes in $S_{I, rel}$. For example, letting $S_G = \cc{\textsc{Urban/rural}}$, $S_{I} = \cc{\textsc{Education}}$, and $S_{I, rel} = \cc{\textsc{Employment}}$, we have queries of the form:
\begin{itemize}
    \setlength\itemsep{0em}
    \item ($\gQ_R$) What proportion of \textit{individuals} who live in an \textbf{urban} area and have a \textbf{graduate degree} are \textit{married} to an individual who is \textbf{employed}?
\end{itemize}
Similar to assumptions made in Section \ref{sec:queries} for the more general setting, we assume that neighboring datasets $D'$ differ on a single individual while maintaining the overall household count. We also assume that households in $D'$ are always valid. For example, households without a head cannot exist. In Appendix \ref{appx:acs_case_study}, we describe how the flexibility of \hpd allows us to make minor changes tailored to this problem domain. Using this modification, we also define $f_q$ for $q \in \gQ_R$.

\textbf{Experiments.}
To evaluate \textit{intra-group-relationship} queries $\gQ_R$, we create a separate sample from the ACS data, which we denote as \textsc{ACS (rel.) NY-19}. In this case, we restrict individual types (attribute \textsc{RELATE}) to the values $\textsc{Head, Spouse, Child}$, giving us a maximum group size of $M=6$ (up to $4$ children and $1$ spouse can exist). To add the queries $\gQ_R$, we again have that $|S| = 3$ where $S = S_{G} \cup S_{I} \cup S_{I, rel}$. With the three relation types \textit{spouse/parent/child}, we have a total of $|\gQ_R| = 877590$ queries. Like before, we also have that $|\gQ_G| = |\gQ_I| = 526684$. 

In Figure \ref{fig:acs_relate_max_error}, we evaluate \fixed and \generative on \textsc{ACS (rel.)} using the sets of queries $\gQ = \gQ_R$ (left column) and $\gQ = \gQ_G \cup \gQ_I \cup \gQ_R$ (right column). Although the performance of the methods are close, we again observe that \fixed performs better.

\input{figures/max_error/acs_relate}

%% file: figures/max_error/acs_relate.tex
\begin{figure}[!tbh]
    \centering
    \includegraphics[width=0.8\linewidth]{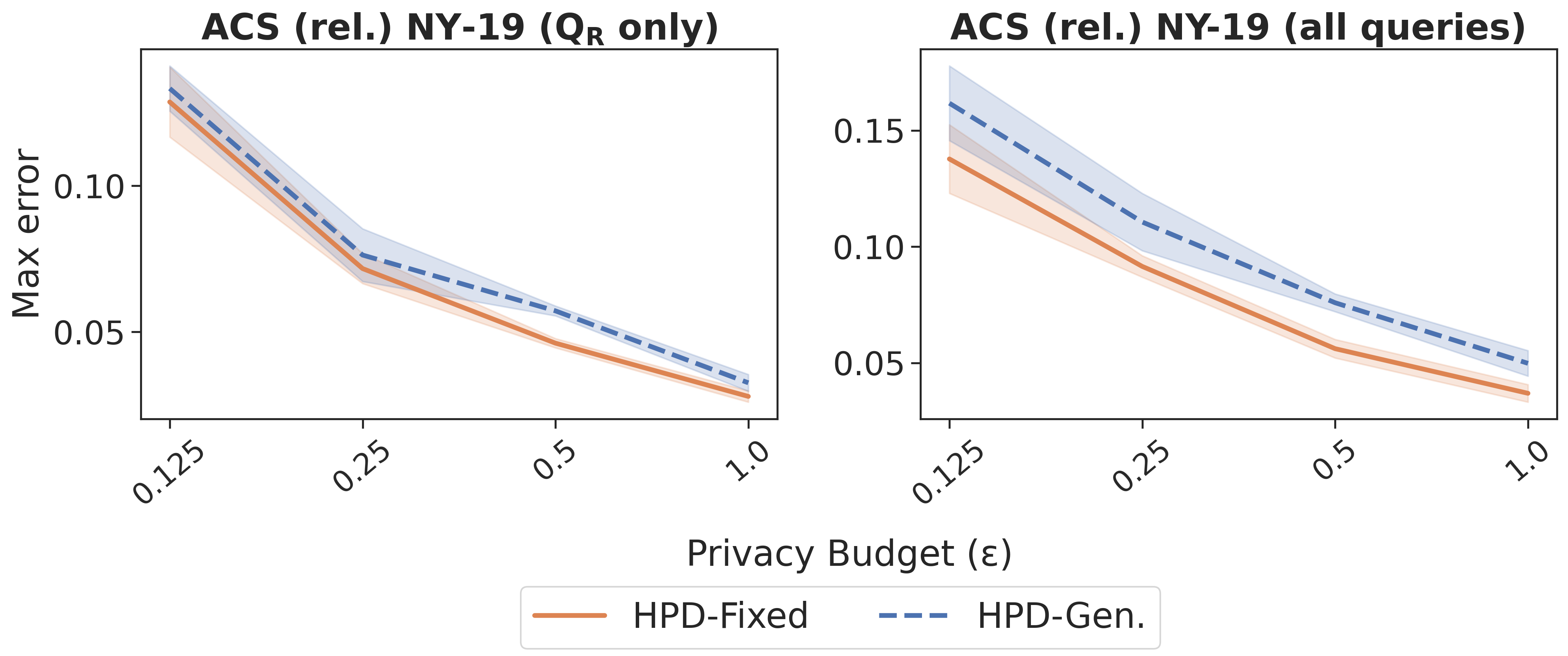}
    \caption{Max errors for queries evaluated on ACS (rel.) NY-19 where $\varepsilon \in \cc{0.125, 0.25, 0.5, 1}$ and $\delta = \frac{1}{N_I^2}$. On the left, we evaluate only on queries in $\gQ_R$, and on the right, we use all queries $\gQ_G \cup \gQ_I \cup \gQ_R$. The \textit{x-axis} uses a logarithmic scale. Results are averaged over $5$ runs, and error bars represent one standard error.
    }
    \label{fig:acs_relate_max_error}
\end{figure}

%% file: docs/conclusion.tex
\section{Conclusion}

In summary, having formulated the problem of hierarchical query release, we present an approach for generating private synthetic data with hierarchical structure. We then empirically evaluate our methods using a variety of query types and datasets, demonstrating that hierarchical relationships can be preserved using private synthetic data. Going forward, we hope that our work will inspire future research on hierarchical data with respect to both synthetic data and private query release.

%% file: docs/appendix.tex
\newpage
\appendix


\input{docs/appendix/algorithms}

\input{docs/appendix/experiments}

%% file: docs/appendix/algorithms.tex
\section{Hierarchical query release methods}\label{appx:algorithms}

We provide additional information regarding the methods discussed in Section \ref{sec:modeling}. Note that implementation details for \mwem, \fixed, and \generative can be found in Appendix \ref{appx:experiments}.


\subsection{Adaptive Measurements}\label{appx:adaptive}

\input{algos/adaptive_full}

For the purposes of making this work more self-contained, we restate details of the \adaptive framework given by \citet{liu2021iterative}. Specifically, we provide the full \adaptive algorithm in Algorithm \ref{alg:adaptive_full}, in which we write down the algorithm using zero concentrated differential privacy (zCDP) \cite{DworkR16,BunS16}
\begin{definition}(Zero Concentrated Differential Privacy \citep{BunS16})\label{def:zCDP}
A randomized mechanism $\gM:\gX^n \rightarrow \mathbb{R}$ satisfies $\rho$-zero concentrated differential privacy ($\rho$-zCDP) if for all neighboring datasets $D, D'$ (i.e., differing on a single person), and for all $\alpha \in \pp{0,\infty}$,
\begin{align*}
    \mathrm{R}_{\alpha}\pp{\gM(D) \parallel \gM(D')} \leq \rho\alpha
\end{align*}
where $\mathrm{R}_{\alpha}\pp{\gM(D) \parallel \gM(D')}$ is the $\alpha$-R\'{e}nyi divergence between the distributions $\gM(D)$ and $\gM(D')$.
\end{definition}

We note that we can convert from zCDP to $(\varepsilon, \delta)$-DP using the following:
\begin{lemma}{\citep{BunS16}}
For all $\delta > 0$, if $\gM$ is $\rho$-zCDP, then $\gM$ satisfies $\pp{\varepsilon, \delta}$-differential privacy where $\varepsilon = \rho + 2\sqrt{\rho \log\pp{1 / \delta}}$.
\label{lem:zcdp_adp}
\end{lemma}

Using this lemma, we can now prove Theorem \ref{thm:privacy}, which states the privacy guarantee of \adaptive in terms of $\pp{\varepsilon, \delta}$-DP. We restate the proof sketch given by \citet{liu2021iterative}.
 
\noindent\textit{Proof sketch of Theorem \ref{thm:privacy}}. 
At each iteration, \adaptive runs the \textit{exponential} and \textit{Gaussian} mechanisms, which satisfy $\frac{1}{2} \pp{\alpha \varepsilon_0}^2$ \citep{cesar2020unifying} and $\frac{1}{2} \sbrack{(1 - \alpha) \varepsilon_0}^2$-zCDP \citep{BunS16} respectively. Therefore at each iteration, \adaptive satisfies $\frac{1}{2} \sbrack{\alpha^2 + (1 - \alpha)^2}\varepsilon_0^2$-zCDP, or $\frac{T}{2} \sbrack{\alpha^2 + (1 - \alpha)^2}\varepsilon_0^2$-zCDP after $T$ iterations \citep{BunS16}. Plugging in $\varepsilon_0 = \sqrt{\frac{2\rho}{T \pp{\alpha^2 + \pp{1-\alpha}^2}}}$, we have that \adaptive satisfies $\rho$-zCDP. By Lemma \ref{lem:zcdp_adp}, \adaptive then satisfies $\pp{\varepsilon, \delta}$-differential privacy where $\varepsilon = \rho + 2\sqrt{\rho \log\pp{1 / \delta}}$. Therefore, for all $\epsilon, \delta > 0$, there exists some $\rho$ such that when run with the \textit{exponential} and \textit{Gaussian} mechanisms using parameters $\alpha \varepsilon_0$ and $(1 - \alpha) \varepsilon_0$ respectively, \adaptive satisfies $\pp{\varepsilon, \delta}$-differential privacy.


\subsection{Explicit distribution over elements in the data universe}

We consider algorithms that optimize over the distributional family $\gD = \cc{\rvx \mid \rvx \in [0, 1]^d, \norm{\rvx}_1 = 1}$. In other words, $\gD$ contains all possible probability distributions over $\gX$, and any histogram $A \in \gD$ can be thought of as a normalized histogram over group types in $\gX$. In this case, the parameters of methods optimizing over such representations are simply the values of a $|\gX|$-dimensional probability vector. Furthermore, we can also sample directly from $A$ to generate a synthetic dataset $\hat{D} \in \gX$.

Suppose we have some dataset $D \in \gX$ with some group size $N_G$ and histogram representation $A_D \in \gD$. Then, we can write any \linear (Definition \ref{def:linear_query}) as a dot product $q_{\phi}(D) = N_G \langle \vec{q}_\phi, A_D \rangle$ and $\vec{q}_\phi = \lbrack \phi(\rvx_1), \ldots, \phi(\rvx_d) \rbrack$. We now discuss how to evaluate both \textit{group} and \textit{individual-level} queries on a histogram $A \in \gD$.

\textit{Group-level queries} ($\gQ_G$).
Given that $A$ is a distribution over all possible group types, we have that $\phi:\gX \rightarrow \cc{0, 1}$. Moreover, since we are normalizing over some total number of groups $N_G$ for \textit{group}-level queries, our query function $f_q$ for some query with predicate $\phi$ is simply
\begin{align*}
    f_{q}(A) = \frac{1}{N_G} q_{\phi}(A) = \langle \vec{q}_\phi, A \rangle
\end{align*}

\textit{Individual-level queries} ($\gQ_I$).
A group can contribute up to $M$ individuals when counting the number of individuals satisfying some predicate condition $\phi:\gX \rightarrow \cc{0, 1, \ldots, M}$. Moreover, even when $N_G$ is fixed, the number of individual rows contained in $D$ varies with $A_D$. Let $\phi_{\#}:\gX \rightarrow \cc{1, 2, \ldots, M}$ be the predicate function that outputs the size of a group $x \in \gX$. Then the total number of individual rows in $D$ is $N_I = N_G \langle \vec{q}_{\phi_{\#}}, A_D \rangle$. Therefore for \textit{individual}-level queries, we have that
\begin{align*}
    f_{q}(A) = \frac{1}{N_I} q_{\phi}(A) = \frac{ \langle \vec{q}_{\phi}, A \rangle}{\langle \vec{q}_{\phi_{\#}}, A \rangle}
\end{align*}

Having described how to both parameterize $\gX$ and evaluate any hierarchical query $q$ on some dataset $D \in \gX$, one can directly optimize our objective using algorithms such as \mwem \citep{hardt2010simple}, which we describe in Appendix \ref{appx:experiments_mwem}, or \pep \citep{liu2021iterative}.


\subsection{Modifications to \hpd for modeling intra-household relationships in the ACS}\label{appx:acs_case_study}

Given the emphasis on the individual type in the ACS, we propose the following modification to $\gX$:
\begin{equation}
    \gX_{\textrm{ACS}} = \gG \times 
    \gI_\textrm{h}\times \tilde{\gI_\textrm{s}} \times \tilde{\gI_\textrm{c}}^{M_c}
\end{equation}
where $\gI_h$, $\gI_s$, and $\gI_c$ correspond to the attributes of individuals with types \textsc{Head}, \textsc{Spouse}, and \textsc{Child} respectively, and $M_h = 1$, $M_s = 1$, $M_c=4$ denotes the maximum number of each type of person in the dataset.\footnote{In the ACS, every household has exactly $1$ \textsc{Head} and up to $1$ \textsc{Spouse}.}

One immediate advantage of this formulation of $\gX_{\textrm{ACS}}$ is that we can now enforce counts for individuals types in our synthetic dataset. Previously, for example, synthetic households sampled from the product distribution $P$ could contain more spouses or children than is possible given the constraints of the ACS data. Similarly, a synthetic household could be entirely missing a person designated as the head.

In our product distribution representation, we also replace $P_{\textrm{\#}}$ with $P_{\textrm{\#}_{h}}$, $P_{\textrm{\#}_{s}}$, and $P_{\textrm{\#}_{c}}$ (i.e., \# head, spouse, and children) and $P_{\gI, i}$ with $P_{\gI_{h}}$, $P_{\gI_{s}}$, and $P_{\gI_{c, i}}$ for $i \in \cc{1, 2, \ldots, M_c}$. Note that because every household in the ACS must have a head, we have that $P_{\textrm{\#}_{h}} = \anglebrack{0, 1}$ is fixed (i.e., the probability of a household having a single head is always $1$). As in Section \ref{sec:modeling}, we will use the following matrix notation: $\vect{P}^{\textrm{\#}_h}, \vect{P}^{\textrm{\#}_s}, \vect{P}^{\textrm{\#}_c}$ and $\vect{P}^{\gI_h}, \vect{P}^{\gI_s}, \vect{P}^{\gI_c, i}$ for $i \in \cc{1, 2, \ldots, M_c}$. 

Finally, we can write down the query function $f_q$ for \textit{intra-group-relationship} queries ($\gQ_R$) as
\begin{align*}
    f_{q} \pp{\vect{P}} 
    = 
    \sum_{k=1}^K
    & \pp{\prod_{i \in S_G} \vect{P}^{\gG}_{ki}}
    \sum_{(\textrm{ref, rel}) \in R_q}
    \pp{
    \sum_{m=1}^{M_\textrm{ref}} \vect{P}^{\textrm{\#}_{\textrm{ref}}}_{km} \sum_{j=1}^{m} \pp{\prod_{i \in S_{I}} \vect{P}^{\gI_{\textrm{ref}, j}}_{ki}}
    } 
    \\ & \times
    \sum_{m=1}^{M_\textrm{rel}} \vect{P}^{\textrm{\#}_{\textrm{rel}}}_{km}
    \pp{1 - \prod_{j=1}^{m}{\pp{1 - \prod_{i \in S_{I,\textrm{rel}}} \vect{P}^{\gI_{\textrm{rel}, j}}_{ki}}}}
    \left. \middle/
    \sum_{k=1}^K \pp{
    1 + 
    \vect{P}^{\textrm{\#}_{\textrm{s}}}_{1m} + 
    \sum_{m=1}^{M_c} m \vect{P}^{\textrm{\#}}_{km}
    } \right.
\end{align*}
where $R_q$ denotes the pairs of person types for each relationship type. For example, for a query $q$ counting the number of people \textit{married} to an individual satisfying some condition, then $R_q = \cc{(h, s), (s, h)}$ (i.e., counting pairs of heads and spouses). We include in Table \ref{tab:intra_relation_queries} more details about \textit{intra-group-relationship} queries that we consider for the ACS dataset. Note that in this equation for $f_q$, $\vect{P}^{\gI_{h, 1}}$ and $\vect{P}^{\gI_{s, 1}}$ just correspond to $\vect{P}^{\gI_h}$ and $\vect{P}^{\gI_s}$ since there can be at most $1$ head and spouse in this dataset.

\input{tables/intra_relation_queries}

%% file: algos/adaptive_full.tex
\begin{algorithm}[!tbh]
\caption{\adaptive}
\label{alg:adaptive_full}

\begin{algorithmic}
\STATE \textbf{Input:} Private dataset $D$, set of queries $\gQ$, distributional family $\gD$, loss function $\gL$
\STATE \textbf{Parameters:} Privacy parameter $\rho$, number of iterations $T$, privacy weighting parameter $\alpha$

\STATE Let $\varepsilon_0 = \sqrt{\frac{2\rho}{T \pp{\alpha^2 + \pp{1-\alpha}^2}}}$
\STATE Let $\Delta = \max_{q \in \gQ} \Delta_1 \pp{q} $ be the max $\ell_1$-sensitivity over all queries in $\gQ$
\STATE Initialize distribution $D_0 \in \gD$

\FOR{$t = 1 $ {\bfseries to} $T$}
\STATE \textbf{Sample}: Choose $\tilde{q}_{t} \in \gQ$ using the \textit{exponential mechanism} with score function
\begin{equation*}
    P[\tilde{q}_t = q] \propto \exp \cc{\frac{\alpha \varepsilon_0}{2 \Delta} |q(D) - q(D_{t-1})}
\end{equation*}
\STATE \textbf{Measure:} Take measurement via the \textit{Gaussian mechanism}:
\begin{equation*}
    m_t = \tilde{q}_t(D) + \gN \pp{0, \pp{\frac{\Delta}{(1 - \alpha) \varepsilon_0}}^2}
\end{equation*}
\STATE \textbf{Update:}
Let $\tilde{Q}_{t} = \bigcup_{i=1}^t \tilde{q}_{t}$ and $\tilde{A}_{t} = \bigcup_{i=1}^t \tilde{a}_{t}$. Update distribution $D$:
\begin{align*}
    D_{t} \leftarrow \argmin_{D \in \gD} \gL \pp{D_{t-1}, \tilde{Q}_{t}, \tilde{A}_{t}}
\end{align*}
\ENDFOR

\STATE \textbf{Output:} $H\pp{\cc{D_t}_{t=0}^{T}}$  where $H$ is some function over all distributions $D_t$ (such as the average)
\end{algorithmic}
\end{algorithm}

%% file: tables/intra_relation_queries.tex
\begin{table}[!bt]
\caption{We describe the different types of household relationships in the ACS (spouse/parent/child) that we consider for queries $\gQ_R$. In particular, for each relationship condition, we list its sensitivity and the set of individual type pairings, $R_q$.}
\begin{center}
\renewcommand{\arraystretch}{1.3}
\begin{tabular}{c | c c c}
\toprule
Relation & Sensitivity & $R_q$ & Example \\
\midrule
\multirow{2}{*}{\textit{married to}} & \multirow{2}{*}{\textit{$\frac{2}{N_I}$}} & \multirow{2}{*}{\textit{$\cc{(h, s), (s, h)}$}} & How many people are \textbf{married to} \\
& & & someone who has graduated college? \\
\multirow{2}{*}{\textit{has child}} & \multirow{2}{*}{\textit{$\frac{2}{N_I}$}} & \multirow{2}{*}{\textit{$\cc{(h, c), (s, c)}$}} & How many people \textbf{have a child} \\
& & & who is in high school? \\
\multirow{2}{*}{\textit{has parent}} & \multirow{2}{*}{\textit{$\frac{M_c}{N_I}$}} & \multirow{2}{*}{\textit{$\cc{(c, h), (c, s)}$}} & How many people \textbf{have a parent} \\
& & & who is older than thirty-years-old? \\
\bottomrule
\end{tabular}
\label{tab:intra_relation_queries}
\end{center}
\end{table}


%% file: docs/appendix/experiments.tex
\section{Experimental Details}\label{appx:experiments}

We provide additional information related to our empirical evaluation. 

\subsection{Mean error results}

We plot mean error for our experiments presented in Section \ref{sec:experiments} in Figures \ref{fig:acs_all_errors} and \ref{fig:acs_relate_all_errors}. Results with respect to mean error are similar to those with respect to max error. We observe, however, that \generative outperforms \fixed on \textsc{ACS (rel.)} when run on $\gQ_R$ only.

\clearpage

\input{figures/all_errors/acs}

\input{figures/all_errors/acs_relate}

\subsection{Data} 

In Table \ref{tab:data_attrs}, we list attributes from the ACS and AFST data that we use for our experiments. We obtained the raw data from the IPUMS USA database \citep{ruggles2021ipums}. The AFST data is confidential, and so access to the raw data currently requires signing a data agreement. We obtained permission from the Allegheny County Office of Children, Youth and Families to publish error plots on experiments run on a sample derived from this data. More information about the attributes used for the tool can be found in \citet{vaithianathan2017developing}.

\input{tables/data_attrs}

\subsection{Algorithm implementation details}\label{appx:mwem}

We provide the exact details of our algorithms in the following sections. Hyperparameters can be found in Table \ref{tab:hyperparameters}. All experiments are run using a desktop computer with an Intel® Core™ i5-4690K processor and NVIDIA GeForce GTX 1080 Ti graphics card. Our implementation is derived from \url{https://github.com/terranceliu/dp-query-release} \citep{liu2021iterative}.

\input{tables/hyperparameters}

\subsubsection{MWEM}\label{appx:experiments_mwem}

\input{algos/mwem}

We restate \mwem in Algorithm \ref{alg:mwem}, with a slight change to the \textit{multiplicative weights} update rule---in our case, we rescale $f_{q_t}(x)$ by a factor of $\frac{1}{M}$ (Algorithm \ref{alg:mwemupdate}) when $q_t$ is an \textit{individual}-level query so that $|f_{q_t}| \le 1$. In addition, we add empirical improvements described in \citet{liu2021leveraging}, which are presented in Algorithm \ref{alg:mwemupdate}. 

Note that in general, running \mwem is extremely impractical in the hierarchical setting since $|\gX|$ scales exponentially with $M$ (in addition to $k_G$ and $k_I)$. For example, in the non-hierarchical setting, the domain size of the $8$ attributes used in \textsc{ACS (red.) NY-19} is only $960$ (compared to a domain size of $262080$ in our case when $M=3$).
\subsubsection{Hierarchical Product Distributions}\label{appx:experiments_hpd}

\input{algos/hpd}

As noted in Section \ref{sec:modeling}, \fixed and \generative share the same training procedure as \gem under the \adaptive framework. For the readers' convenience, we include this procedure in Algorithms \ref{alg:hpd} and \ref{alg:hpdupdate}. We note that in our experiments using \generative, $F$ is a multi-headed MLP with two hidden layers that have residual connections between each.

As presented in Section \ref{sec:modeling}, \hpd maintains a separate probability matrix $\vect{P}^{\gI, i}$ for each possible person in $\gX$. However, one can further reduce the computational requirements by reusing some probability matrix $\vect{P}^{\gI}$ for multiple rows in each group. For example, at the other extreme, one could potentially have just a single matrix $\vect{P}^{\gI, i}$ for all $m$ rows in a group, significantly reducing the number of parameters of the model at the cost of expressivity (i.e., characteristics of individual rows are now completely independent of the group size itself). In our standard experiments, we follow a scheme $\cc{1, 1, 1, 2, 2, 3}$, meaning that the first three individuals in a household are sampled from $\vect{P}^{\gI_{1}}$, $\vect{P}^{\gI_{2}}$, and $\vect{P}^{\gI_{3}}$ respectively, the next two from $\vect{P}^{\gI_{4}}$ etc. Similarly for \textsc{ACS (rel.)}, we use a single matrix $\vect{P}^{\gI_{\textrm{c}}}$ for all $M_c$ children (i.e., all children in a household are drawn from the same distribution). For future work, it may be interesting to assess how well this strategy can scale to values of $M$ that are several times larger than that of our datasets.

Finally, \citet{liu2021iterative} show that running their algorithm, \gem, without resampling Gaussian noise $\vect{z}$ at each gradient step significantly speeds up training time while not harming performance. However, we find that in this setting, our method \generative performs much better when the noise is fixed, as shown in Figure \ref{fig:acs_compare}. Regardless, however, we find that \fixed performs slightly better.

\input{figures/acs_compare}

%% file: figures/all_errors/acs.tex
\begin{figure}[!ht]
    \centering
    \includegraphics[width=0.9\linewidth]{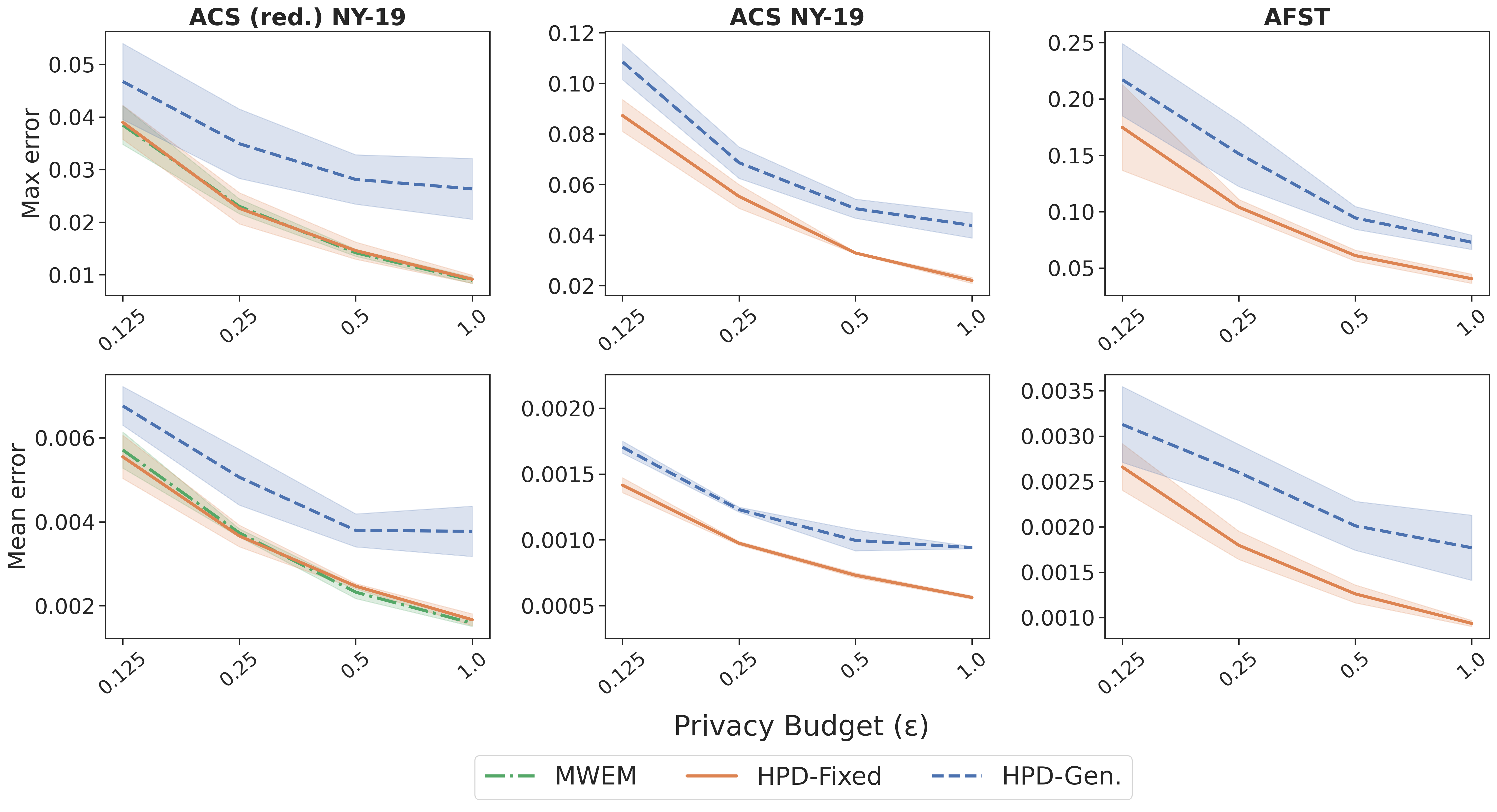}
    \caption{Max and mean errors for \textit{group} and \textit{individual}-level hierarchical queries evaluated on ACS/ACS (red.) NY-19 and AFST where $\varepsilon \in \cc{0.125, 0.25, 0.5, 1}$ and $\delta = \frac{1}{N_I^2}$. The \textit{x-axis} uses a logarithmic scale. Results are averaged over $5$ runs, and error bars represent one standard error.
    }
    \label{fig:acs_all_errors}
\end{figure}

%% file: figures/all_errors/acs_relate.tex
\begin{figure}[!ht]
    \centering
    \includegraphics[width=0.7\linewidth]{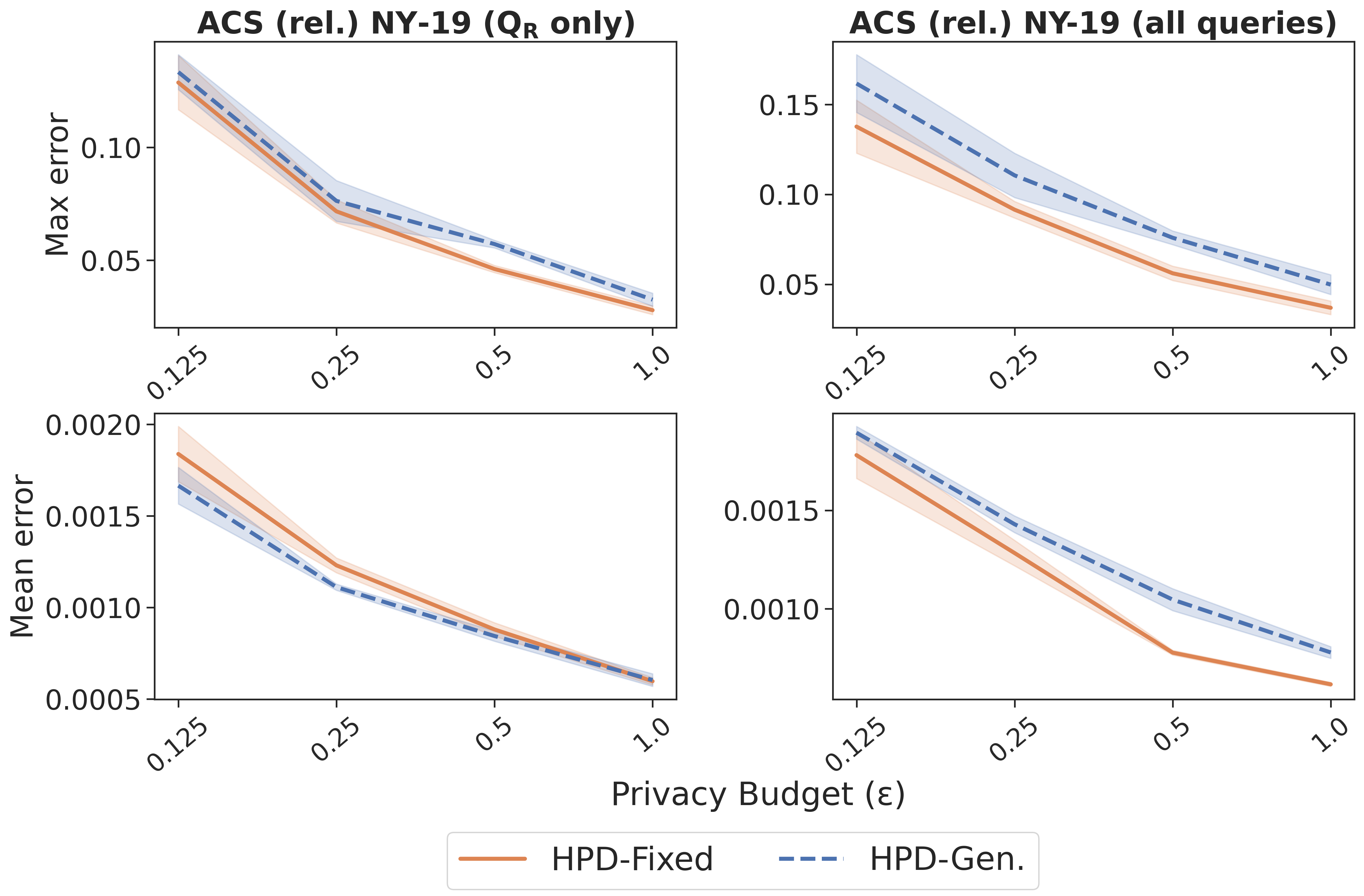}
    \caption{Max and mean errors for queries evaluated on ACS (rel.) NY-19 where $\varepsilon \in \cc{0.125, 0.25, 0.5, 1}$ and $\delta = \frac{1}{N_I^2}$. On the left, we evaluate only on queries in $\gQ_R$, and on the right, we use all queries $\gQ_G \cup \gQ_I \cup \gQ_R$. The \textit{x-axis} uses a logarithmic scale. Results are averaged over $5$ runs, and error bars represent one standard error.
    }
    \label{fig:acs_relate_all_errors}
\end{figure}

%% file: tables/data_attrs.tex
\begin{table}[!ht]
\renewcommand{\arraystretch}{1.3}
\centering
\caption{Data Attributes. Note that all hierarchical datasets have an attribute \textsc{COUNT} that denotes the group size.}
\label{tab:data_attrs}
\begin{tabular}{c  c | c}
    \toprule
    Dataset & Domain & Attributes \\
    \midrule
    \multirow{2}{*}{\textsc{ACS (red.)}}
    & \multirow{1}{*}{$\gG$} 
    & \textsc{COUNT, METRO, OWNERSHP, FARM, FOODSTMP} \\ \cmidrule(lr){2-3}
    & \multirow{1}{*}{$\gI$} 
    & \textsc{SEX, AGE, EMPST, MARST} \\
    \midrule
    \multirow{6}{*}{\textsc{ACS / ACS (rel.)}}
    & \multirow{3}{*}{$\gG$} 
    & \textsc{COUNT, METRO, OWNERSHP, FARM, COUNTYFIP} \\ 
    & & \textsc{FARMPROD, ACREHOUS, ROOMS} \\
    & & \textsc{BUILTYR2, FOODSTMP, MULTGEN} \\ \cmidrule(lr){2-3}
    & \multirow{3}{*}{$\gI$} 
    & \textsc{SFRELATE, SEX, MARST, RACE} \\ 
    & & \textsc{HISPAN, CITIZEN, EDUC, SCHOOL} \\
    & & \textsc{EMPSTAT, LOOKING, AGE} \\
    \midrule
    \multirow{9}{*}{\textsc{AFST}}
    & \multirow{3}{*}{$\gG$} 
    & \textsc{VERSION, RELATIONSHIP\_TO\_REPORT} \\
    & & \textsc{CALL\_SCRN\_OUTCOME, SERVICE\_DECISION} \\
    & & \textsc{HH\_CITY} \\ \cmidrule(lr){2-3}
    & \multirow{6}{*}{$\gI$}
    & \textsc{ACJ\_NOW\_VICT\_SELF, ACJ\_EVERIN\_VICT\_SELF} \\
    & & \textsc{JPO\_NOW\_VICT\_SELF, JPO\_EVERIN\_VICT\_SELF} \\ 
    & & \textsc{AGE\_AT\_RFRL\_VICT\_SELF} \\
    & & \textsc{PLSM\_PAST548\_COUNT\_NULL} \\
    & & \textsc{SER\_PAST548\_COUNT\_VICT\_SELF} \\ 
    & & \textsc{REF\_PAST548\_COUNT\_VICT\_SELF} \\
    \bottomrule
\end{tabular}
\end{table}

%% file: tables/hyperparameters.tex
\begin{table}[!ht]
\centering
\caption{Hyperparameters. We use $\alpha = 0.67$ for all methods.}
\label{tab:hyperparameters}
\renewcommand{\arraystretch}{1.3}
\begin{tabular}{l l c c}
    \toprule
    Dataset & Method & Parameter & Values \\
    \midrule
    \multirow{7}{*}{\textsc{ACS (red.)}} & 
    \multirow{2}{*}{\mwem} & 
    \multirow{2}{*}{$T$} & $100$, $200$, $300$, $400$ \\
    & & & $500$, $750$, $1000$ \\ \cmidrule(lr){2-4}
    & \multirow{2}{*}{\fixed} & learning rate & $0.1$ \\
    & & $T$ & $50$, $75$, $100$, $125$, $150$, $200$, $250$ \\ \cmidrule(lr){2-4}
    & \multirow{3}{*}{\generative} & learning rate & $0.0001$ \\
    & & hidden layers & $(512, 1024)$ \\
    & & $T$ & $50$, $75$, $100$, $125$, $150$, $200$, $250$ \\
    \midrule
    \multirow{5}{*}{\textsc{ACS / ACS (rel.) / AFST}} & \multirow{2}{*}{\fixed} & learning rate & $0.1$ \\
    & & $T$ & $100$, $200$, $300$, $400$, $500$ \\ \cmidrule(lr){2-4}
    & \multirow{3}{*}{\generative} & learning rate & $0.0001$ \\
    & & hidden layers & $(512, 1024)$ \\
    & & $T$ & $100$, $200$, $300$, $400$, $500$ \\
    \bottomrule
\end{tabular}
\end{table}

%% file: algos/mwem.tex
\begin{algorithm}[!tbh]
\caption{\mwem}
\label{alg:mwem}
\begin{algorithmic}
\STATE {\bfseries Input:} Private hierarchical dataset $D \in \gX$, set of queries $Q$
\STATE {\bfseries Parameters:} Privacy parameter $\rho > 0$, number of iterations $T$, privacy weighting parameter $\alpha$, max per-round iterations $T_{\text{max}}$

\STATE Let $M$ be the maximum possible number of individual rows belonging to a group in $\gX$
\STATE Let $\Delta = \max_{q \in Q} \Delta_1 \pp{q} $ be the max $\ell_1$-sensitivity over all queries in $Q$
\STATE Let $\varepsilon_0 = \sqrt{\frac{2\rho}{T \pp{\alpha^2 + \pp{1 - \alpha}^2}}}$
\STATE Let $A_D$ be a the normalized histogram representation of $D$
\STATE Initialize $A_0$ be a uniform distribution over $\gX$

\FOR{$t = 1 $ {\bfseries to} $T$}
\STATE \textbf{Sample:} Select query $\tilde{q}_t \in Q$ using the \emph{exponential mechanism} with score function
\begin{equation*}
    P[\tilde{q}_t = q] \propto \exp \cc{\frac{\alpha \varepsilon_0}{2 \Delta} |f_{q}(A_D) - f_{q}(A_{i-1})}
\end{equation*}
\STATE \textbf{Measure:} Take measurement via the \textit{Gaussian mechanism}
\begin{equation*}
    m_t = f_{\tilde{q}_t}(A_D) + \gN \pp{0, \pp{\frac{\Delta}{(1 - \alpha) \varepsilon_0}}^2}
\end{equation*}
\STATE \textbf{Update:} $A_t$ = \mwemupdate($A_{t-1}, \tilde{Q}_{t}, \tilde{M}_{t}, T_\text{max}$) where $\tilde{Q}_{t} = \anglebrack{\tilde{q}_1, \ldots, \tilde{q}_t}$ and $\tilde{M}_{t} = \anglebrack{\tilde{m}_1, \ldots, \tilde{m}_t}$
\ENDFOR

\STATE \textbf{Output:} $A = \text{avg}_{t \in [T]} A_{t-1}$
\end{algorithmic}
\end{algorithm}

\begin{algorithm}[!tbh]
\caption{\mwemupdate}
\label{alg:mwemupdate}
\begin{algorithmic}
\STATE {\bfseries Input:} Normalized histogram $A$, queries $Q = \anglebrack{q_1, \ldots, q_t}$, noisy measurements $M = \anglebrack{m_1, \ldots, m_t}$, max iterations $T_\text{max}$ \\

\STATE Let $a_\text{max} = \max_{1 \le t \le |Q|} |m_t - f_{q_t}(A)|$ be the max error across queries in $Q$ \\
\STATE Let $S_{\text{top}}$ be the collection of indices $t$ for the top $T_\text{max}$ queries with highest error $|m_t - f_{q_t}(A)|$ \\
\STATE Let $S_\text{threshold} = \cc{t \mid t \in S_{\text{top}}, |m_t - f_{q_t}(A)| \ge  \frac{a_\text{max}}{2}}$ be the collection of indices $t \in S_\text{threshold}$ such that the error for $q_t$ is greater than $\frac{a_\text{max}}{2}$\\

\FOR{$t \in \randomize(S_\text{threshold})$}
\STATE Let $A$ be a distribution s.t.
\begin{equation*}
    A(x) \propto A(x)\exp \cc{f_{\hat{q_t}}(x) \pp{\frac{m_t - f_{q_t}(A)}{2}}}
\end{equation*}
where
\begin{equation*}
    f_{\hat{q}}(x) = \begin{cases} 
      f_{q}(x) & q \in \gQ_G \\
      \frac{1}{M} f_{q}(x) & q \in \gQ_I
  \end{cases}
\end{equation*}
\ENDFOR

\STATE \textbf{Output:} $A$
\end{algorithmic}
\end{algorithm}


%% file: algos/hpd.tex
\begin{algorithm}[!tbh]
\caption{\hpd}
\label{alg:hpd}
\begin{algorithmic}
\STATE {\bfseries Input:} Private hierarchical dataset $D \in \gX$, queries $Q$
\STATE {\bfseries Parameters:} Privacy parameter $\rho > 0$, number of iterations $T$, privacy weighting parameter $\alpha$, batch size $B$, max per-round iterations $T_{\text{max}}$

\STATE Let $\Delta = \max_{q \in Q} \Delta_1 \pp{q} $ be the max $\ell_1$-sensitivity over all queries in $Q$
\STATE Let $\varepsilon_0 = \sqrt{\frac{2\rho}{T \pp{\alpha^2 + \pp{1 - \alpha}^2}}}$
\STATE Initialize $\vect{P}_0$ as some representation of a hierarchical product distribution that is parameterized by $\theta$ (either fixed table in \fixed or neural network in \generative)

\FOR{$t = 1 $ {\bfseries to} $T$}
\STATE \textbf{Sample:} 
Select query $\tilde{q}_t \in Q$ using the \emph{exponential mechanism} with parameter $\varepsilon_0$ and score function
\begin{equation*}
    P[\tilde{q}_t = q] \propto \exp \cc{\frac{\alpha \varepsilon_0}{2 \Delta} |q(D) - q(A_{i-1})}
\end{equation*}
\STATE \textbf{Measure:} Take (via the \textit{Gaussian mechanism}) measurement 
\begin{equation*}
    m_t = \tilde{q}_t(D) + \gN \pp{0, \pp{\frac{\Delta}{(1 - \alpha) \varepsilon_0}}^2}
\end{equation*}
\STATE \textbf{Update:} $F_t$ = \hpdupdate($F_{t-1}, \tilde{Q}_{t}, \tilde{M}_{t}, T_\text{max}, \gamma$) where $\tilde{Q}_{t} = \anglebrack{\tilde{q}_1, \ldots, \tilde{q}_t}$, $\tilde{M}_{t} = \anglebrack{\tilde{m}_1, \ldots, \tilde{m}_t}$, and $\gamma = \ema\pp{|\tilde{Q}_{t} - \tilde{M}_{t}|}$
\ENDFOR

\STATE Let $\theta_{out} = \ema \pp{\cc{\theta_j}_{j=\frac{T}{2}}^T}$ where $\theta_j$ parameterizes $F_j$ \\
\STATE Let $F_{out}$ be the generator parameterized by $\theta_{out}$ \\
\STATE \textbf{Output:} $F_{out} \pp{\vect{z}}$ \\
\end{algorithmic}
\end{algorithm}

\begin{algorithm}[!tbh]
\caption{\hpdupdate}
\label{alg:hpdupdate}
\begin{algorithmic}
\STATE {\bfseries Input:} $\vect{P}$, queries $Q = \anglebrack{q_1, \ldots, q_t}$, noisy measurements $M = \anglebrack{m_1, \ldots, m_t}$, max iterations $T_\text{max}$, stopping threshold $\gamma$ \\

\STATE Sample $\vect{z} = \anglebrack{z_1 \ldots z_B} \sim \mathcal{N}(0, I_B)$ \\
\STATE Let $\vect{a} = M - f_Q(\vect{P})$ be errors over queries in $Q$ (where $f_{Q}(\cdot) = \anglebrack{f_{q_1}(\cdot), \ldots, f_{q_t}(\cdot)}$) \\

\FOR{$t = 1 $ {\bfseries to} $T$}
\STATE Let $S_{\text{threshold}} = \cc{i \mid |a_i| \ge \gamma}$
\STATE Let $\hat{Q} = \cc{q_i \mid i \in S_{\text{threshold}}}$ and $\hat{M} = \cc{m_i \mid i \in S_{\text{threshold}}}$
\STATE Update $\vect{P}$ via stochastic gradient
decent according to Equation \ref{eq:hpd_loss}: $\gL^{\hpd} \pp{\hat{Q}, \hat{M}}$
\STATE Resample $\vect{z} = \anglebrack{z_1 \ldots z_B} \sim \mathcal{N}(0, I_B)$ and update $\vect{a} = M - f_Q(\vect{P})$ \\
\IF{$\max_i |a_i| < \gamma$}
\STATE \textbf{break};
\ENDIF
\ENDFOR
\STATE \textbf{Output:} $F$
\end{algorithmic}
\end{algorithm}

%% file: figures/acs_compare.tex
\begin{figure}[!ht]
    \centering
    \includegraphics[width=0.9\linewidth]{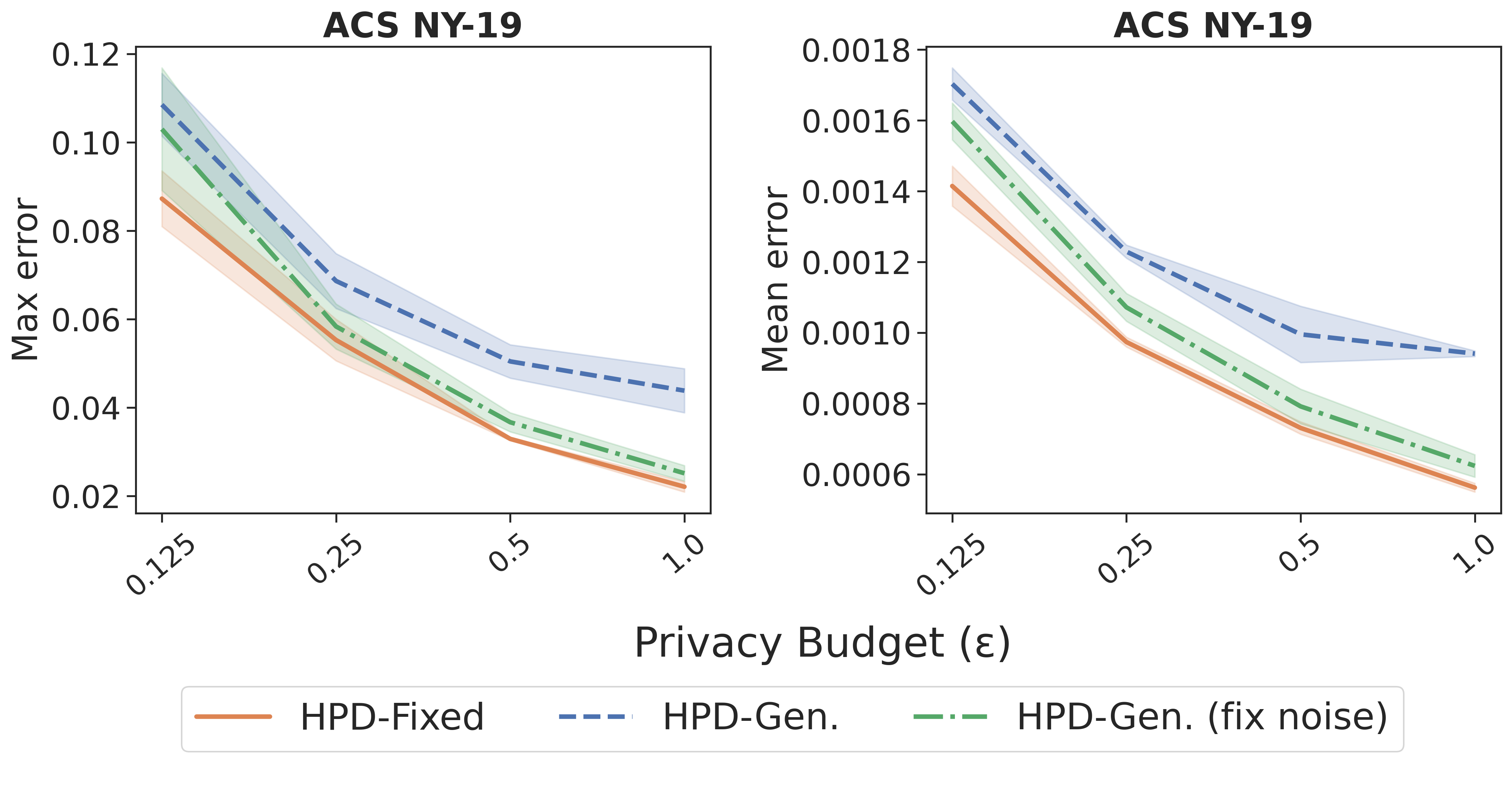}
    \caption{We compare running \generative without resampling Gaussian noise $\vect{z}$ at each gradient step (HPD-Gen.(fix noise)) to \generative with resampling (HPD-Gen.). We plot max and mean errors for \textit{group} and \textit{individual}-level hierarchical queries evaluated on ACS NY-19 where $\varepsilon \in \cc{0.125, 0.25, 0.5, 1}$ and $\delta = \frac{1}{N_I^2}$. The \textit{x-axis} uses a logarithmic scale. Results are averaged over $5$ runs, and error bars represent one standard error.
    }
    \label{fig:acs_compare}
\end{figure}